\documentclass[preprint,12pt]{elsarticle}

\usepackage{lipsum}
\usepackage{ntheorem}
\usepackage{mathtools}
\usepackage{url}
\usepackage{amsmath}

\newtheorem*{problem-non}{Problem}
\usepackage{color, soul}
\usepackage{comment}
\usepackage{array}
\usepackage{multirow}

\newcolumntype{L}[1]{>{\raggedright\let\newline\\\arraybackslash\hspace{0pt}}m{#1}}
\newcolumntype{C}[1]{>{\centering\let\newline\\\arraybackslash\hspace{0pt}}m{#1}}
\newcolumntype{R}[1]{>{\raggedleft\let\newline\\\arraybackslash\hspace{0pt}}m{#1}}

\newenvironment{tightcenter}{%
  \setlength\topsep{0pt}
  \setlength\parskip{0pt}
  \begin{center}
}{%
  \end{center}
}

\usepackage{graphicx}
\usepackage{amssymb}

\journal{Journal Name}

\begin{document}

\begin{frontmatter}


\title{Named Entity Sequence Classification}



\author{Mahdi Namazifar}

\address{Twitter Inc.}

\begin{abstract}
Named Entity Recognition (NER) aims at locating and classifying named entities in text. 
In some use cases of NER, including cases where detected named entities are used in creating content recommendations, it is crucial to have a reliable confidence level for the detected named entities.  In this work we study the problem of finding confidence levels for detected named entities. We refer to this problem as Named Entity Sequence Classification (NESC).  We frame NESC as a binary classification problem and we use NER as well as recurrent neural networks to find the probability of candidate named entity is a real named entity.  We apply this approach to Tweet texts and we show how we could find named entities with high confidence levels from Tweets.

\end{abstract}


\end{frontmatter}


\section{Introduction}
\label{S:1}

In Named Entity Recognition (NER) the goal is to locate and classify named entities (defined as physical or abstract objects that can be expressed with proper nouns) in a given text.  NER as an area of Natural Language Processing (NLP) has been studied extensively. A survey of studies of NER based on classical NLP approaches can be found in  \cite{jbp:/content/journals/10.1075/li.30.1.03nad}.  NER approaches based on deep learning have also been frequently reported in the NLP literature \cite{DBLP:journals/corr/SantosG15, Socher:2012:DLN:2390500.2390505}.  These approaches generally rely on pre-trained word embeddings as well as sequence modeling techniques based on Recurrent Neural Networks (RNNs) or Temporal Convolutions.

NER for Tweets has also been studied in numerous studies \cite{Ritter:2011:NER:2145432.2145595, Liu:2011:RNE:2002472.2002519, Liu:2013:NER:2414425.2414428, Limsopatham2016BidirectionalLF}. Due to the limit on the number of characters of a Tweet, heavy use of slangs and emojis, lack of proper capitalization, as well as the informal style of writing, detecting named entities in Tweet text is significantly more challenging than other types of text (News, books, web page, etc.).  

Our focus in this work is on detecting named entities in Tweets.  The architecture that we use for our NER model is somewhat similar to the architecture proposed in \cite{DBLP:journals/corr/HuangXY15} in that we use bidirectional LSTMs and Conditional Random Fields (CRF) to tag Tweets tokens with named entity labels.  We show that our trained NER model performs quite well relative to the standard open source NER solution by the Stanford NLP group \cite{manning-EtAl:2014:P14-5}.

One of the applications of NER on Tweets for Twitter is recommending content to users. Some example content recommendation use cases where NER is valuable include notifying the user that their network is Tweeting about a specific named entity, or grouping Tweets about a certain named entity to show the users. It is important to note that for recommending content it is crucial to have very high confidence in the detected named entities to be actually true named entities.  Unfortunately it is not straightforward to measure confidence on a sequence of tokens that is detected by our NER model (which is a sequence tagging based approach) being a named entity.  In fact none of the studies that we found on sequence tagging would address this requirement, and that is the motivation behind the problem that we call Named Entity Sequence Classification (NESC).

We define NESC as follows: given a text and a sequence of tokens in that text, determine if the sequence of tokens is a named entity.  The idea is that NESC as a binary classification problem would provide the probability of the sequence of tokens (named entity candidate) being in fact a named entity.  For the NESC model we use the output of the trained bidirectional LSTM of the NER model for a context window around the candidate named entity (sequence of tokens that we want to classify as a named entity or not) and we train another LSTM for the binary classification problem of NESC.

It should be mentioned that this binary classification of sub-sequences of a sequence is not unique to named entity recognition and could be applied to any sequence tagging problem where consecutive elements of the sequence could constitute a tag of interest.

In the rest of the this paper, we first present our NER model for Tweets. Next we discuss the NESC problem definition, our proposed model architecture for NESC, and also details on how to train the proposed NESC model. Finally we show some experimental results on performance of our NER and NESC models.

\section{NER for Tweets}
By definition, NER not only tries to locate named entities in text, but also tries to classify the detected named entities to a pre-determined set of entity types. Different NER models have different sets of possible entity types.  In this work for the set of possible entity types that are covered by our NER model, we adhered to the following types: \textit{Person}, \textit{Place}, \textit{Product}, \textit{Organization}, and \textit{Other}; and we use the Inside--Outside--Beginning (IOB) format \cite{IOB} for entity boundaries.

In order to build an NER model for Tweets we follow the general model architecture proposed by many authors \cite{DBLP:journals/corr/TranMJ17, DBLP:journals/corr/ChiuN15, Limsopatham2016BidirectionalLF} that includes pre-trained word embeddings for Tweet tokens followed by a recurrent neural network (a bidirectional LSTM in our case) and a fully dense layer followed by a softmax layer that produce a discrete probability distribution for the possible labels of each Tweet token. 

\begin{figure}[h]
\centering\includegraphics[width=0.4\linewidth]{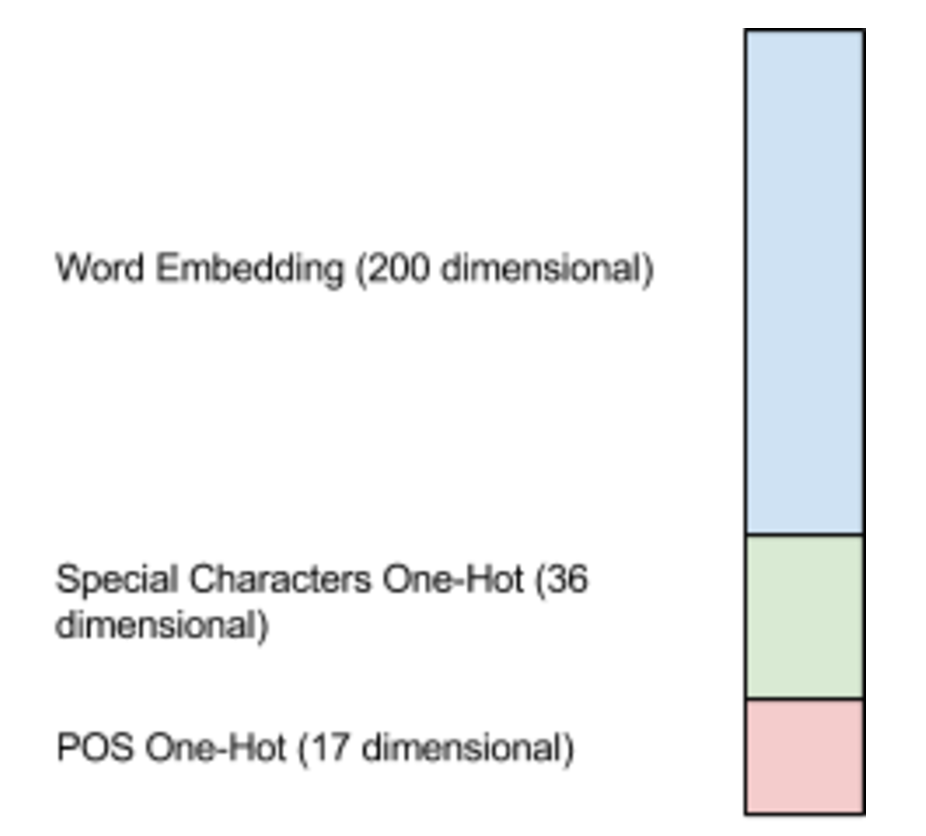}
\caption{Vectorized token}
\label{fig_vectorized_token}
\end{figure}

Tweet texts are first tokenized using a Twitter internal text processing tool which is a heuristic-based text tokenizer for tens of languages.  Each token then is vectorized using our pre-trained word embeddings.  Our word embeddings are 200-dimensional vectors that are trained using GloVe \cite{pennington2014glove} on over 1 billion Tweets.  Other than dense word embeddings, the vector representing each token also includes some sparse variables in the form of 2 one-hot vectors. The first one-hot vector indicates whether the token is one of the special characters (\%, /, ., !, ?, \ldots), a hashtag, an @handle, whether it's first character is capitalized, or the entire token is capitalized \footnote{If the entire token is capitalized, the value of the indicator for first character being capitalized is 0.}. This one-hot vector has 36 dimensions.  The other one-hot vector specifies the Part of Speech (POS) tag associated with each of the tokens. These POS tags are also provided by Twitter's internal text processing tools, and each token can get one of the 17 possible POS tags; and as a result this one-hot vector has 17 dimensions.  For the non-zero value of the one-hot vectors we use 0.1 so that this number is in the same range as the values in the dense word embedding vectors.  Figure \ref{fig_vectorized_token} depicts the parts of a vectorized token. 

\begin{figure}[h]
\label{ner_model}
\centering\includegraphics[width=0.6\linewidth]{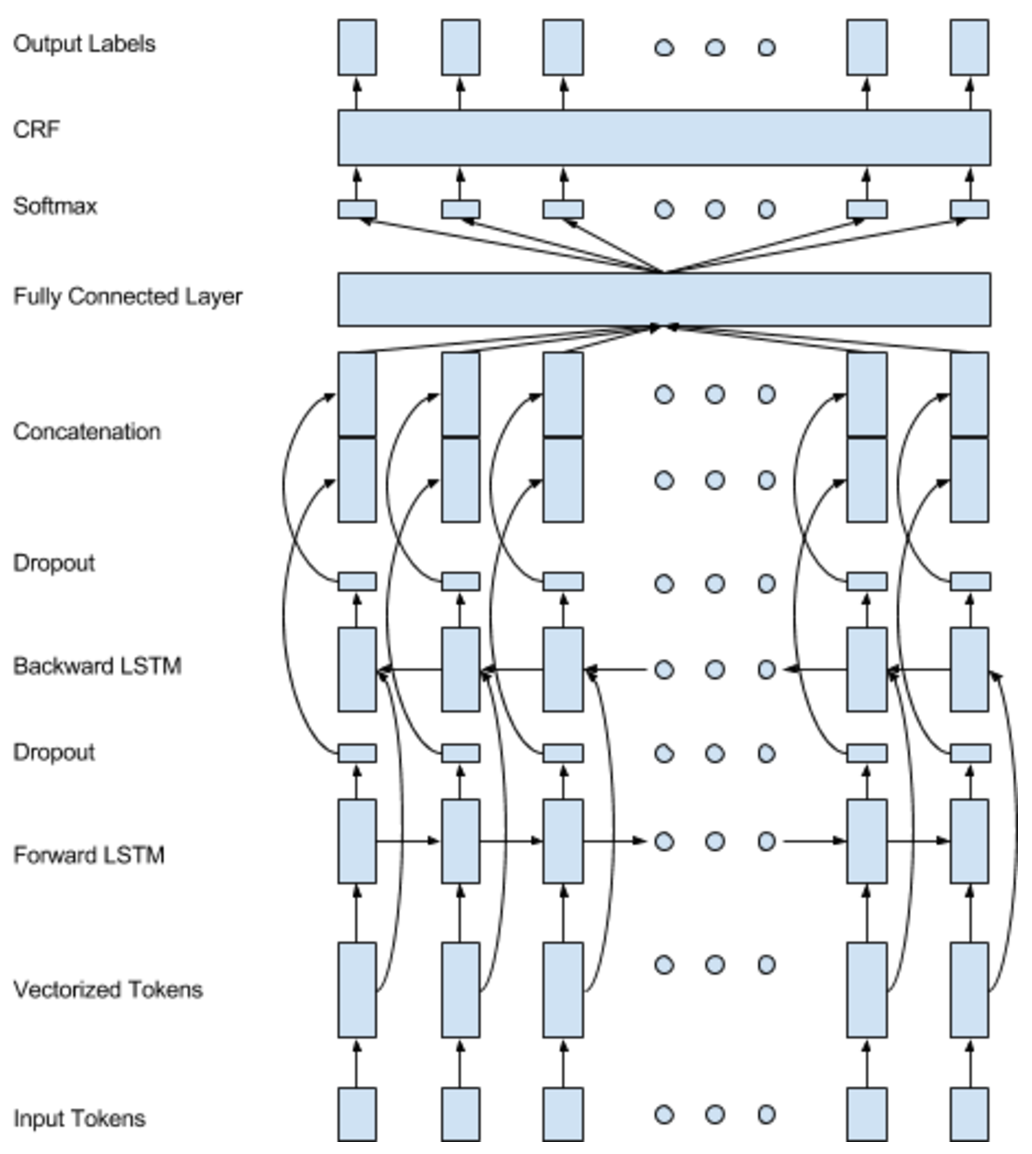}
\caption{NER model architecture}
\end{figure}

Vectorized  Tweet tokens next go through our NER model that has the architecture that is shown in Figure \ref{ner_model}.  As one could see from the figure vectorized tokens go through a bidirectional LSTM with dropout and the output goes through a fully connected layer and then a softmax layer to create 11 dimensional probability distributions for each token.  These 11 dimensions correspond to beginning and inside of the 5 entity type tags plus the one \textit{not-an-entity} tag. Next these probabilities go through a Conditional Random Field (CRF) which learns the correct order of entity labels and the result is the final NER labels for each of the tokens. For instance it learns that \textit{not-an-entity} labels most likely is not followed by a label for inside an entity label (e.g., \textit{I-person}).

\section{NESC}
\subsection{Problem Definition}
The output of the NER model discussed in the previous section for each token provides a label along with a probability that is given by the softmax layer to that label. These scores however cannot directly be used, for example in a multiplicative way, to get a confidence for multi-token entities. As an example if we run sentence \textit{``I love San Francisco"} through the NER model the output of the softmax layer (probability of each label for each token) are shown in Table \ref{ner-example-probabilities}.

\begin{table}[h]
\centering
\begin{tabular}{ l  c  c  c  c }
\hline
\textbf{} & \textbf{I} & \textbf{love} & \textbf{San} & \textbf{Francisco}\\
\hline
O-not-an-entity & \textbf{0.985} & \textbf{0.988} & 0.012 & 0.026\\
B-Person & 0.001 & 0.001 & 0.023 & 0.043\\
I-Person & 0.001 & 0.001 & 0.001 & 0.087\\
B-Place & 0.002 & 0.001 & \textbf{0.483} & 0.005\\
I-Place & 0.001 & 0.001 & 0.006 & \textbf{0.659}\\
B-Product & 0.003 & 0.001 & 0.276 & 0.021\\
I-Product & 0.001 & 0.001 & 0.159 & 0.032\\
B-Organization & 0.001 & 0.002 & 0.009 & 0.123\\
I-Organization & 0.001 & 0.001 & 0.011 & 0.002\\
B-Other & 0.001 & 0.001 & 0.012 & 0.001\\
I-Other & 0.003 & 0.002 & 0.008 &0.001\\
\hline
\end{tabular}
\caption{Example of NER labels and their associated probabilities}
\label{ner-example-probabilities}
\end{table}

 From the labels in Table \ref{ner-example-probabilities} we can see that the model predicts that "San Francisco" is a place, but it is not straightforward to calculate the likelihood of ``San Francisco" being an entity from the softmax probabilities provided for the token labels. The reason for that is that these probabilities are calculated using the entire context of the text and all of the probabilities of tokens and their labels are correlated.
 
 The NESC problem aims at calculating likelihoods for named entities that are proposed by the output of the NER model. We define the NESC problem as follows:
\begin{problem-non}
Given a text $T$ and a sub-sequence of tokens $S$ in $T$, what is the probability of the $S$ being a named entity in $T$.
\end{problem-non} 

For instance in our previous example \textit{``I love San Francisco"} what is the probability of the sequence containing the two tokens \textit{``San"} and \textit{``Francisco"} being a named entity?  In this example 

\begin{align*} 
  T & : \hspace{5mm}\text{I love San Francisco} \\
  S & : \hspace{5mm}\{\text{San}, \text{Francisco}\}
\end{align*}
and $S$ is the output of the NER model.

\subsection{NESC Model Architecture}
\begin{figure}[h]
\label{nesc-context-window}
\centering\includegraphics[width=1\linewidth]{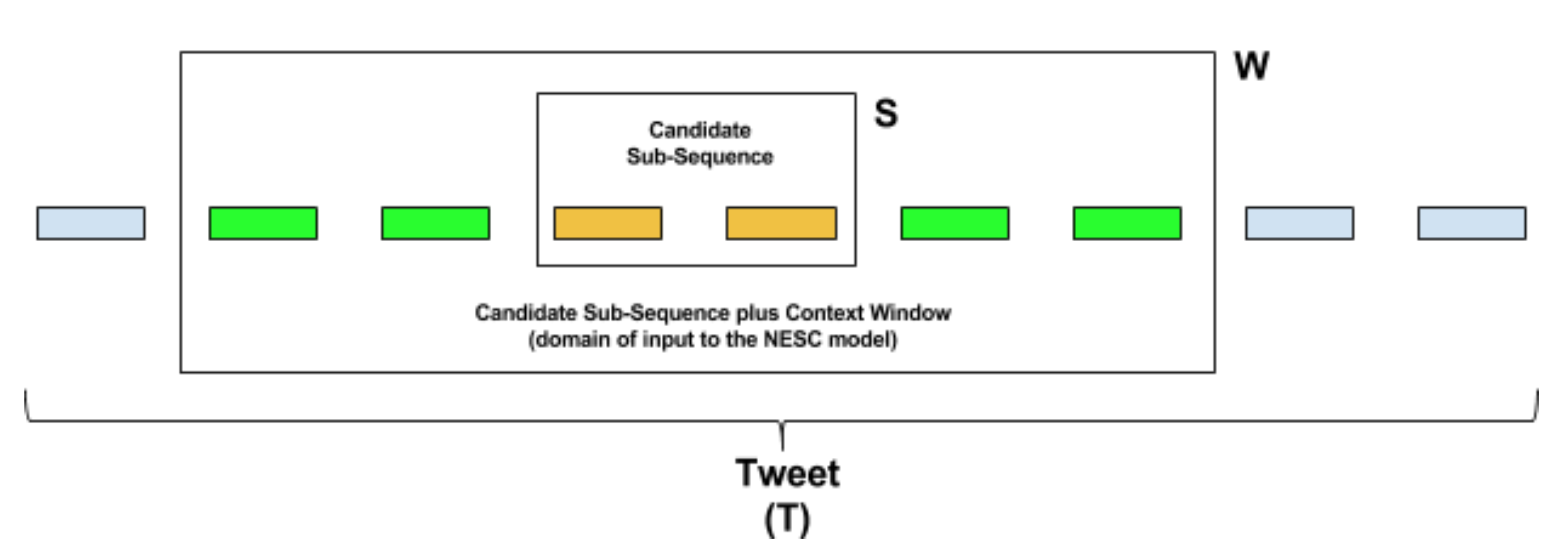}
\caption{NESC context window}
\end{figure}

By definition, NESC is a binary classification problem, and in the rest of this section we discuss how we build a model for this problem.  First we define a context window around $S$ with context size $k$. For instance if the index of the starting token of $S$ is $i$ and the index of the ending token of $S$ is $j$, then the context window of $S$ would be tokens $i-k$ to $j+k$.  Let's call this context window $W$. Needless to say we first pad the input text $T$ on both sides with pad size $k$ to make sure that we can define $W$ for any sub-sequence of tokens in $T$. Figure \ref{nesc-context-window} shows an example of context window around a candidate named entity of size 2 tokens and context size ($k$) of 2.

Now NESC as a binary classification problem boils down to given a context window with context size $k$, whether the middle part of the window (considering the context size $k$) is a named entity or not. To approach this problem we need to vectorize the context window defined by the sequence of the candidate tokens $S$ and the other tokens in the context window.  To do this we use the output of pre-trained NER model's bidirectional LSTM and we build a sequence of the bidirectional LSTM layer output vectors that are associated with the tokens of the context window of $S$.  In other words the vectorization of context window is simply the slice of the output of the pre-trained NER model's bidirectional LSTM layer output that coincides with the context window.

Remember that for each token, the NER model's bidirectional LSTM vector associated with it has captured information from all other tokens that come before and after it, and therefore the vectorization of context window contains information from the entire text $T$ and not only the context window $W$. 

\begin{figure}[h]
\label{nesc-architecture}
\centering\includegraphics[width=0.8\linewidth]{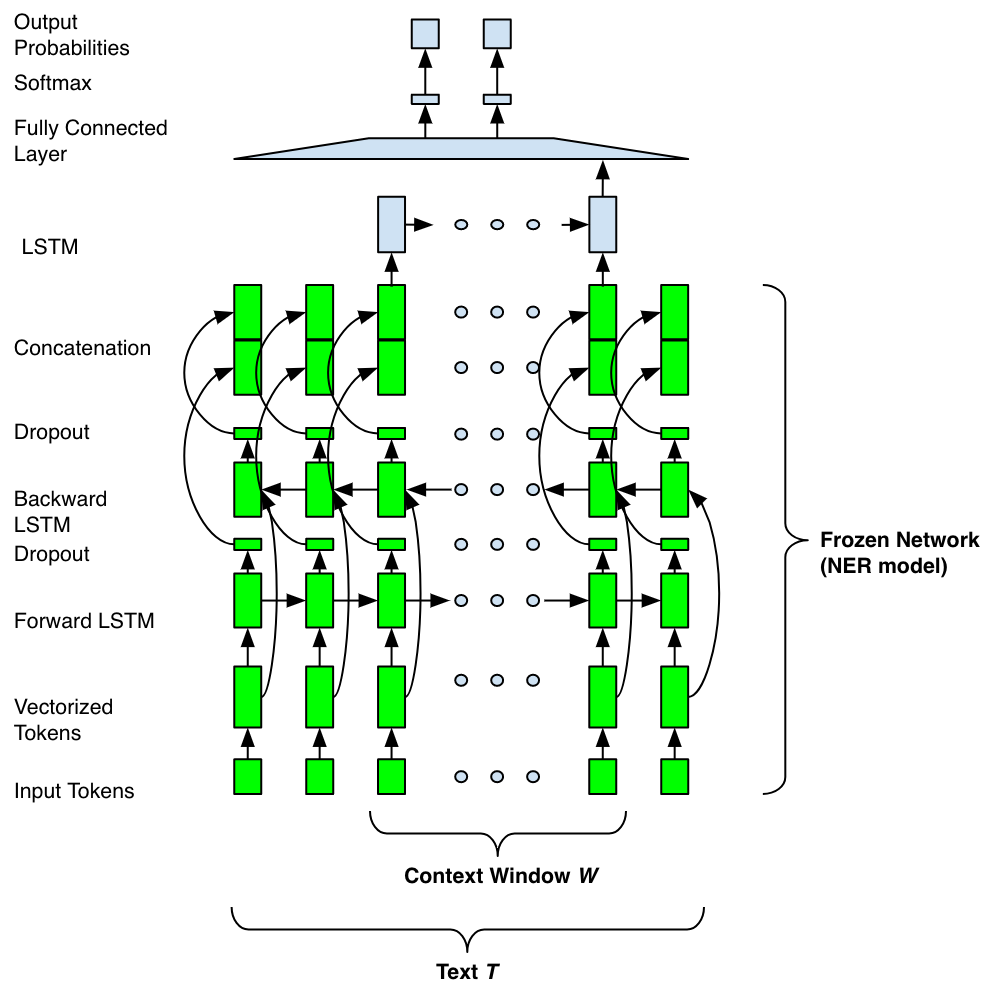}
\caption{NESC model architecture}
\end{figure}

Now that we have the context window $W$ represented as a sequence of vectors, the NESC problem can be viewed as a binary classification problem on sequences. For this we build a model that is an LSTM from which we take the internal state vector and follow that by a fully connected layer and a softmax layer that outputs the probability of the candidate sub-sequence $S$ being a named entity. Figure \ref{nesc-architecture} depicts the full architecture of the NESC model.

\subsection{Model Training}
\label{nesc-data-prep}
We build the training set for NESC from the labeled data of the NER problem. Each NESC training sample is a sequence of vectors that are the output of the bidirectional LSTM layer of the pre-trained NER model for a context window, along with a target binary value which indicates whether the center of the context window (with a given context size) is a named entity or not. For positive samples (i.e., sub-sequences of tokens that are named entities in the text that they appear in), we can directly use named entities in the NER training data.  Each labeled named entity in that data becomes a record in the NESC training set with positive target value. 

On the other hand, the negative samples are generated in two different ways. First way of generating negative samples involves perturbations on positive samples. More specifically, if we consider the window of tokens of a positive sample in the text, we can get negative samples by extending, shrinking, and moving this window.  Table \ref{negative-sample} shows some negative samples that can be created by perturbing the correct named entity.

\begin{table}[h!]
\scriptsize
\centering
\begin{tabular}{l l l}
\hline
\hline
 homeless population in \textit{\textbf{\color{red}San Francisco}} is surging & Positive NESC Sample \\
 homeless population in San \textit{\textbf{\color{red}Francisco}} is surging & Negative NESC Sample \\
 homeless population in \textit{\textbf{\color{red}San}} Francisco is surging & Negative NESC Sample \\
 homeless population \textit{\textbf{\color{red}in San}} Francisco is surging & Negative NESC Sample \\
 homeless population in San \textit{\textbf{\color{red}Francisco is}} surging & Negative NESC Sample \\
 
\hline
\end{tabular}
\caption{Positive and negative samples for NESC from NER labeled data}
\label{negative-sample}
\end{table}
As it is shown for each named entity in the NER labeled data, one positive and several negative NESC samples can be created.  

The second approach of generating negative samples for NESC is simply taking a random sub-sequence of tokens from the NER labeled text.  For each random sub-sequence we also check to make sure that the sub-sequence is not in fact an entity.  This is easy to do since these random sub-sequences are selected from NER labeled texts. For each random sub-sequence we first sample the size of the sub-sequence from the empirical discrete distribution of size (number of tokens) of the named entities in the NER labeled data. Next we select a random sub-sequence of tokens of length that we just sampled. Lastly we check to make sure that the selected random sub-sequence is not a named entity. This can be automatically done by checking the labels of tokens of the random sub-sequence.

Due to the imbalance in the number of positive and negative training samples created, we use a weighted cross entropy loss function for NESC in which the weights of the binary class are calculated based on number of positive and negative samples in the training set.

\subsection{NER Labeled Data}
In order to create the labeled data for NER we first took a sample of Tweets. Twitter's Human Computations Team (HCOMP) labeled the tokens of each  Tweet sample based on the IOB schema.  Each sample was initially labeled by two different individuals and for samples that the two human labels did not match, a third labeler also labeled the samples.  At the end we ended up with around 100,000 labeled English Tweets on which at least 2 labelers completely agreed on the labels.  From this set we created our training, validation, and test sets.

In our NER training set there are 62,507 named entities, each of which would become a positive sample for NESC.  On the other hand, using the approaches discussed earlier we create 226,067 negative NESC samples, and as a result our NESC training set contains 288,574 records.

\section{Results}
We first report the performance of our NER model on our labeled test set and compare it with the performance of Stanford NLP's \cite{manning-EtAl:2014:P14-5} NER model.  Table \ref{ner-stanford-comparison} summarizes this performance comparison.  Here we look at 3 performance measures:
\begin{itemize}
\item \textbf{Untyped Token Level:} The classification problem here is defined on token labels (whether a token gets an entity or a not-an-entity label). A token's label would either associate it with an entity (label starts with ``B-'' or ``I-'') or associate it with not an entity (label O-not-an-entity). In this case relevant instances are tokens that have labels other than O-not-an-entity in the test set. 

\item \textbf{Untyped Entity Level:} The classification problem here is define on named entities (whether an entity is correctly identified). Here named entity types (Person, Place, etc.) are disregarded. In this case relevant instance are full labeled named entities without their types in the test set.  Here the focus is on detecting full named entities and not on their detected type.

\item \textbf{Typed Entity Level:} The classification problem here is defined on typed named entities (whether an entity and its type is correctly identified).  In this case relevant indices are full labeled named entities and their types in the test set.  Here the focus is on detecting full named entities along with their types.
\end{itemize}

\begin{table}[h]
\centering
\scriptsize
\begin{tabular}{l L{3cm} c c c } 
\hline
\textbf{Measure Type} & \textbf{} & \textbf{Precision} & \textbf{Recall} & \textbf{F1 Score} \\
\hline
\multirow{2}{*}{\textbf{Untyped Token Level}}& Twitter NER & 0.84 & 0.78 & 0.81 \\
& Stanford NLP NER & 0.77 & 0.53 & 0.63 \\
\hline
\multirow{2}{*}{\textbf{Untyped Entity Level}}& Twitter NER & 0.76 & 0.71 & 0.73 \\
& Stanford NLP NER & 0.63 & 0.39 & 0.48 \\
\hline
\multirow{2}{*}{\textbf{Typed Entity Level}}& Twitter NER & 0.69 & 0.64 & 0.66 \\
& Stanford NLP NER & 0.54 & 0.34 & 0.42 \\
\hline
\end{tabular}
\caption{Our NER model (Twitter NER) vs Stanford NLP NER model performance on Tweets}
\label{ner-stanford-comparison}
\end{table}

As we can see from these numbers our NER model performs significantly better than Stanford NLP's NER model on Tweets with respect to precision, recall, and F1 score at both typed and untyped entity levels.  This performance difference is especially more apparent in recall numbers where for instance the typed entity recall of our model is 0.64 whereas the same value for Stanford NLP NER is 0.34.

\begin{figure}[h]
\label{pr}
\centering\includegraphics[width=0.8\linewidth]{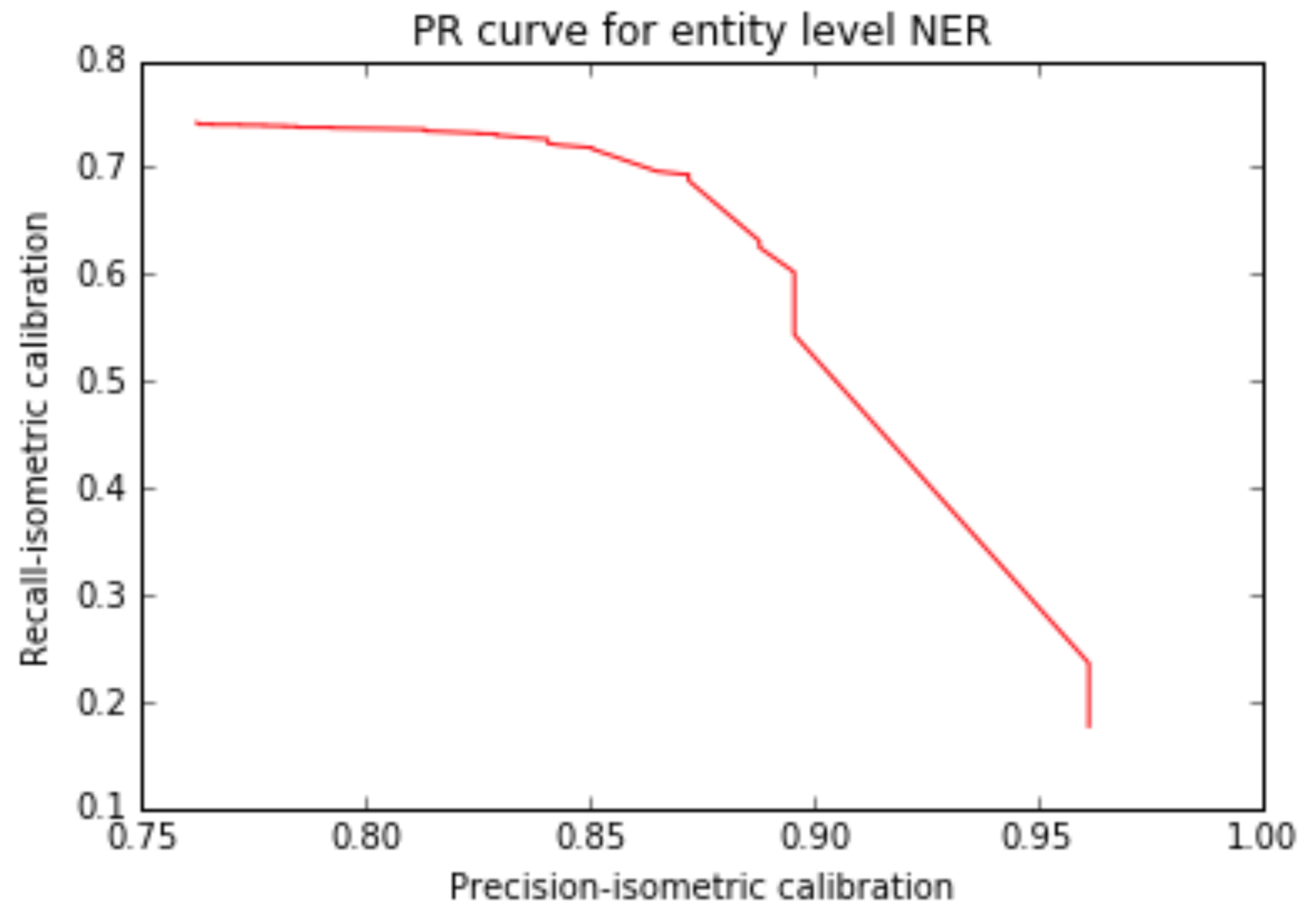}
\caption{NESC Precision--Recall curve}
\end{figure}

Next we study the performance of our NESC model. We created training, validation, and test sets for NESC by applying the method mentioned in Section \ref{nesc-data-prep} on our NER's training, validation, and test sets, respectively. We calculate the precision and recall of the NESC model after isotonic calibration on the validation set at different classification threshold values for the test set. Figure \ref{pr} shows this Precision--Recall curve. 

One could see from this curve that at 0.90 precision the recall is around 0.52 which means that more than half of the entities in the Tweets in the test set are detected with 0.90 precision.  The sparsity in this precision--recall curve is due to the small size of the test set (built using only 2000 Tweets) and could be remedied by getting more labeled Tweets in order to get a higher resolution precision--recall curve.

\begin{table}[h!]
\centering
\scriptsize
\begin{tabular}{ l | L{9cm} | L{3cm}  } 
\hline
 & \textbf{Entities detected by NER with NESC probabilities in tokenized Tweets} & \textbf{Other candidates probability}  \\
\hline
1 & NowPlaying \hl{No Cigarette Smoking In My Room} $_{\color{red}{\text{other }{0.961}}}$ - \hl{Stephen Marley} $_{\color{red}{\text{person }{0.973}}}$ ft \hl{Melonie Fiona} $_{\color{red}{\text{person }{0.985}}}$ https://t.co/mrlTZ0dmx4 16 : 39  &   \\
\hline
2 & @SInow : \hl{Tony Romo} $_{\color{red}{\text{person }{0.985}}}$ reportedly is unlikely to play Sunday vs . the \hl{Steelers} $_{\color{red}{\text{organization }{0.978}}}$ https://t.co/WnWU8zaw4f https://t.co/f4hjMFMkoS & \\
\hline
3 & @GuardianBooks : \hl{The Essex Serpent} $_{\color{red}{\text{other }{0.912}}}$ beats \hl{Harry Potter} $_{\color{red}{\text{person }{0.978}}}$ to win \hl{Waterstones} $_{\color{red}{\text{organization }{0.994}}}$ book of the year https://t.co/hNEQJnRz0p & \\
\hline
4 & \hl{Marco Republic Paris Memory Foam Cushion Womens Mary} $_{\color{red}{\text{product }{0.837}}}$  Jane Platform Wedges Heels Comfort Pumps https://t.co/s1XZ19Oqt4 \#pumps \#flats \#heels & \hl{Marco Republic Paris Memory Foam Cushion Womens Mary Jane Platform Wedges Heels Comfort Pumps} $\color{red}{_{0.925}}$ \\
\hline
5 & See How \hl{Bobrisky} $_{\color{red}{\text{person }{1.000}}}$ And \hl{Lolo} $_{\color{red}{\text{person }{1.000}}}$ Was Dancing On Stage As Fans React [ Video ] https://t.co/PGUTaIztMT https://t.co/Y5zv0Jnjrr & \\
\hline
6 & @90sNiallftafi : when \hl{calum} $_{\color{red}{\text{person }{1.000}}}$ and \hl{Michael} $_{\color{red}{\text{person }{1.000}}}$ got \hl{Ashton} $_{\color{red}{\text{person }{1.000}}}$ to get a spider out of the bathroom \& Ashton scared calum https://t.co/5ieqdZAWTN & \\
\hline
7 & @DaiIyRap : \hl{Childish Gambino} $_{\color{red}{\text{person }{0.978}}}$ is dropping his new album `` \hl{Awaken , My Love} $_{\color{red}{\text{other }{0.994}}}$ '' Next Month https://t.co/qg5a42vnaa & \\
\hline
8 & \hl{Palmetto Packings Compression} $_{\color{red}{\text{product }{0.645}}}$ Seals - 5 / 16 " Square - FDA Listed - 20 Feet - New https://t.co/MmiPthCQNS https://t.co/xsXX4GVO5i & \hl{Palmetto Packings Compression Seals} $\color{red}{_{0.837}}$ \newline \hl{FDA} $\color{red}{_{0.943}}$\\
\hline
9 & First impressions : \hl{Russell Wilson} $_{\color{red}{\text{person }{0.985}}}$ and \hl{Seahawks} $_{\color{red}{\text{organization }{1.000}}}$ sputter on offense in loss to \hl{Bucs} $_{\color{red}{\text{organization }{0.994}}}$ https://t.co/fTyvUduM17 & \\
\hline
10 & Power Forward with us tonight at our \hl{President's Community Lecture} $_{\color{red}{\text{other }{0.925}}}$ featuring \hl{Bill Ritter} $_{\color{red}{\text{person }{0.973}}}$ .  https://t.co/YD1bOqp5bT https://t.co/x5tlAdPGEV & \\
\hline
11 & Happy \hl{Veterans day} $_{\color{red}{\text{other }{0.961}}}$ to every soldier who has fought for the rights of our country @Justin\_\_martin3 love you & \\
\hline
12 & Long \hl{Island Volleyball} $_{\color{red}{\text{organization }{0.148}}}$ College Showcase at \hl{SPORTIME} $_{\color{red}{\text{organization }{1.000}}}$ Thanks to all the Players and College Coaches for making it a huge success ! \#LIVCS https://t.co/7KzLeH8D8a & \hl{Long Island} $\color{red}{_{0.836}}$ \newline \hl{Long Island Volleyball} $\color{red}{_{0.223}}$ \newline \hl{Long Island Volleyball College Showcase}$\color{red}{_{0.285}}$\\ 
\hline
\end{tabular}
\caption{NER and NESC results on sample Tweets}
\label{examples}
\end{table}

Table \ref{examples} shows the result of NER and NESC on a number of sample Tweets. Each row in this table contains a tokenized  Tweet text along with named entities in the text detected by NER (highlighted in yellow), the entity type detected by NER (shown as a subscript in red), and the probability of the detected entity being in fact an entity found by NESC (also shown as a subscript in red). For instance

\begin{tightcenter}
\vspace{5mm}``\hl{Barack Obama} $_{\color{red}{\text{person }{0.993}}}$ is the 44th president of the United States'' 
\end{tightcenter}\vspace{5mm}
indicates that NER has detected that ``Barack Obama'' is a person's name because NER's label for ``Barack'' is B--Person, for ``Obama'' is I--Person, and for ``is'' is O--not--and--entity. Moreover, according to NESC the probability of the substring ``Barack Obama'' in the string ``Barack Obama is the 44th president of the United States'' being an entity is 0.993.

In row 1 of the Table \ref{examples} we see that NER correctly has detected the song name ``No Cigarette Smoking In My Room'' as entity of type other and NESC gives the probability of 0.961 to this entity.  Note that this is a rather long name with 6 tokens which NER has correctly identified. Also there are 2 other named entities (persons) in the  Tweet that are correctly identified by NER and receive high NESC scores.  Row 4 shows a  Tweet in which a NER does not identify a product name correctly and consequently the NESC score that the incorrect subsequence gets is relatively lower than the NESC score for the correct entity (last column).  In row 5 we see a Tweet text in which all the words are capitalized, which is quite common in Tweets. In this example 2 persons names are correctly identified by NER and both of them receive a score of 1.0 from NESC. In row 6 we see a lower case named entity ``calum'' is correctly identified by NER and receive very high scores from NESC. In Row 8 we see that NER incorrectly returns the sequence ``Palmetto Packings Compression'' as a product name where in fact the correct product name is ``Palmetto Packings Compression Seals''.  However it is worth noting that the NESC score for the the incorrect product name is quite low at 0.645, where as if we query the correct product name from NESC the returned score is 0.837 (column 3). Moreover in this example NER fails to detect ``FDA'' as an entity, but if we query NESC for FDA in that Tweet, the returned score is quite high at 0.943. In row 11 we  can see that NER correctly identifies ``Veterans day'' as an entity of type other which has a high NESC score of 0.961. Finally in row 12 we see that NER incorrectly identifies ``Island Volleyball'' as an organization, but we see that the NESC score of that subsequence is very low at 0.148.

For our application of recommending users named entities from Tweets, from these results we see that we can rely on NESC scores with a high threshold.  Specially if for a collection of Tweets NESC gives very high scores (say all 0.95 or higher) to a given sequence of tokens, then we could confidently assume that said subsequence of tokens in fact is an entity, and build user recommendations based on that entity accordingly.




\section{References}
\bibliographystyle{model1-num-names}







\end{document}